\documentclass[11pt,a4paper]{article}
\usepackage[hyperref]{naaclhlt2018}
\usepackage{times}
\usepackage{latexsym}

\usepackage{url}

\usepackage{times}  
\usepackage{helvet}  
\usepackage{courier}  
\usepackage{url}  
\usepackage{graphicx}  
\usepackage{latexsym}
\usepackage{amsmath}
\usepackage{verbatim}
\usepackage{tikz-qtree}
\usepackage{tikz}

\usetikzlibrary{shapes.geometric}
\usetikzlibrary{shapes.arrows}
\usetikzlibrary{patterns}
\usetikzlibrary{chains}
\usetikzlibrary{calc}

\aclfinalcopy 

\title{Multimodal Emoji Prediction}

\author{Francesco Barbieri$^{\diamondsuit}$ ~~ Miguel Ballesteros$^{\spadesuit}$ ~~ \textbf{Francesco Ronzano$^{\heartsuit}$ ~~ Horacio Saggion$^{\diamondsuit}$}\\
$^\diamondsuit$ Large Scale Text Understanding Systems Lab, TALN, UPF, Barcelona, Spain \\
$^\spadesuit$IBM Research, U.S \\
$^\heartsuit$ Integrative Biomedical Informatics Group, GRIB, IMIM-UPF, Barcelona, Spain \\\\
{ \tt $^\diamondsuit$$^\heartsuit$\{name.surname\}@upf.edu}, 
{ \tt$^\spadesuit$miguel.ballesteros@ibm.com}
}

\date{}

\begin{document}
\maketitle
\begin{abstract}
Emojis are small images that are commonly included in social media text messages. The combination of visual and textual content in the same message builds up a modern way of communication, that automatic systems are not used to deal with. 
In this paper we extend recent advances in emoji prediction by putting forward a multimodal approach that is able to predict emojis in Instagram posts. Instagram posts are composed of pictures together with texts which  sometimes include emojis. We show that these emojis can be predicted by using the text, but also using the picture. Our main finding is that incorporating the two synergistic modalities, in a combined model, improves accuracy in an emoji prediction task. This result demonstrates that these two modalities (text and images) encode different information on the use of emojis and therefore  can complement each other.
\end{abstract}

\section{Introduction}

In the past few years the use of emojis in social media has increased exponentially, changing the way we communicate. 
The combination of visual and textual content poses new challenges for information systems which need not only to deal with the semantics of text but also that of images. 
Recent work \cite{barbieri2017emojis} has shown that textual information can be used to predict emojis associated to text. In this paper we show that in the current context of multimodal communication where texts and images are combined in social networks, visual information should be combined with texts in order to obtain more accurate emoji-prediction models. 

We explore the use of emojis in the social media platform Instagram. 
We put forward a multimodal approach to predict the emojis associated to an Instagram post, given its picture and text\footnote{In this paper we only utilize the first comment issued by the user who posted the picture.}.
Our task and experimental framework are similar to \cite{barbieri2017emojis}, however, we use different data (Instagram instead of Twitter) 
 and, in addition, we rely on images to improve the selection of the most likely emojis to associate to a post. We show that a multimodal approach (textual and visual content of the posts) increases the emoji prediction accuracy compared to the one that only uses textual information. This suggests that textual and visual content embed different but complementary features of the use of emojis.

In general, an effective approach to predict the emoji to be associated to a piece of content may help to improve natural language processing tasks \cite{novak2015sentiment}, such as information retrieval, generation of emoji-enriched social media content, suggestion of emojis when writing text messages or sharing pictures online. Given that emojis may also mislead humans \cite{miller2017understanding}, the automated prediction of emojis may help to achieve better language understanding. As a consequence, by modeling the semantics of emojis, we can improve highly-subjective tasks like sentiment analysis, emotion recognition and irony detection \cite{felbo2017using}.

\section{Dataset and Task}
\label{sec:dataset}

\textbf{Dataset:}
We gathered Instagram posts published between July 2016 and October 2016, and geo-localized in the United States of America. We considered only posts that contained a photo together with the related user description of at least 4 words and exactly one emoji.


Moreover, as done by \newcite{barbieri2017emojis}, we considered only the posts which include {\em one and only one} of the 20 most frequent emojis (the most frequent emojis are shown in Table \ref{tab:20detailed}). Our dataset is composed of 299,809 posts, each containing a picture, the text associated to it and only one emoji. In the experiments we also considered the subsets of the 10 (238,646 posts) and 5 most frequent emojis (184,044 posts) (similarly to the approach followed by \newcite{barbieri2017emojis}).

\noindent\textbf{Task}: We extend the experimental scheme of \newcite{barbieri2017emojis}, by considering also visual information when modeling posts. We cast the emoji prediction problem as a classification task: given an image or a text (or both inputs in the multimodal scenario) we select the most likely emoji that could be added to (thus used to label) such contents. 
The task for our machine learning models is, given the visual and textual content of a post, to predict the single emoji that appears in the input comment.

\section{Models}
\label{sec:models}
We present and motivate the models that we use to predict an emoji given an Instagram post composed by a picture and the associated comment.

\subsection{ResNets}
\label{sec:resnet}
Deep Residual Networks (ResNets) \cite{he2016deep} are Convolutional Neural Networks which were competitive in several image classification tasks \cite{ILSVRC15,lin2014microsoft} and showed to be one of the best CNN architectures for image recognition. ResNet is a feed-forward CNN that exploits ``residual learning'', by 
bypassing two or more convolution layers (like similar previous approaches \cite{sermanet2011traffic}). 
We use an implementation of the original ResNet where the scale and aspect ratio augmentation are from \cite{szegedy2015going}, the photometric distortions from \cite{howard2013some} and weight decay is applied to all weights and biases (instead of only weights of the convolution layers).
The network we used is composed of 101 layers (ResNet-101), initialized with pretrained parameters learned on ImageNet \cite{imagenet_cvpr09}. We use this model as a starting point to later finetune it on our emoji classification task.  Learning rate was set to 0.0001 and we early stopped the training when there was not improving in the validation set.

\subsection{FastText}
\label{sec:fasttext}
Fastext
\cite{joulin2016bag} is a linear model for text classification. We decided to employ FastText as it has been shown that on specific classification tasks, it can achieve competitive results, comparable to complex neural classifiers (RNNs and CNNs), while being much faster. FastText represents a valid approach when dealing with social media content classification, where huge amounts of data needs to be processed and new and relevant information is continuously generated.
The FastText algorithm is similar to the CBOW algorithm \cite{mikolov2013efficient}, where the middle word is replaced by the label, in our case the emoji. Given a set of $N$ documents, the loss that the model attempts to minimize is the negative log-likelihood over the labels (in our case, the emojis):
\begin{align*}
loss = -\frac{1}{N}\sum_{N}^{n=1}e_n\log (\mathit{softmax}(BA_{x_n}))
\end{align*}
\noindent where $e_n$ is the emoji included in the $n$-th Instagram post, represented as hot vector, and used as label. A and B are affine transformations (weight matrices), and $x_n$ is the unit vector of the bag of features of the $n$-th document (comment). The bag of features is the average of the input words, represented as vectors with a look-up table.

\subsection{B-LSTM Baseline}
\label{blstm}
\newcite{barbieri2017emojis} propose a recurrent neural network approach for the emoji prediction task. We use this model as baseline, to verify whether FastText achieves comparable performance. They used a Bidirectional LSTM with character representation of the words \cite{ling:2015,lstmemnlp15} to handle orthographic variants (or even spelling errors) of the same word that occur in social media (e.g. \emph{cooooool} vs \emph{cool}).  

\section{Experiments and Evaluation}
\label{sec:exp}
In order to study the relation between Instagram posts and emojis, we performed two different experiments. In the first experiment (Section~\ref{sec:comparison}) we compare the FastText model with the state of the art on emoji classification (B-LSTM) by \newcite{barbieri2017emojis}. Our second experiment (Section~\ref{sec:mm}) evaluates the visual (ResNet) and textual (FastText) models on the emoji prediction task. Moreover, we evaluate a multimodal combination of both models respectively based on visual and textual inputs. Finally we discuss the contribution of each modality to the prediction task.

We use 80\% of our dataset (introduced in Section~\ref{sec:dataset}) for training, 10\% to tune our models, and 10\% for testing (selecting the sets randomly).


\subsection{Feature Extraction and Classifier}
\label{features}
To model visual features we first finetune the ResNet (process described in Section~\ref{sec:resnet}) on the emoji prediction task, then extract the vectors from the input of the last fully connected layer (before the softmax). The textual embeddings are the bag of features shown in Section~\ref{sec:fasttext} (the $x_n$ vectors), extracted after training the FastText model on the emoji prediction task.

With respect to the combination of textual and visual modalities, we adopt a middle fusion approach \cite{kiela2015multi}: we associate to each Instagram post a multimodal embedding obtained by concatenating the unimodal representations of the same post (i.e. the visual and textual embeddings), previously learned. 
Then, we feed a classifier\footnote{L2 regularized logistic regression} with visual (ResNet), textual (FastText), or multimodal feature embeddings, and test the accuracy of the three systems.

\subsection{B-LSTM / FastText Comparison}
\label{sec:comparison}
\begin{table}
\centering
\setlength{\tabcolsep}{3.5pt} 
\renewcommand{\arraystretch}{1} 
\scalebox{0.87}{
\begin{tabular}{|r|ccc|ccc|ccc|}
\hline 
 & \multicolumn{3}{c|}{\textbf{top-5}} 
 & \multicolumn{3}{c|}{\textbf{top-10}}
 & \multicolumn{3}{c|}{\textbf{top-20}} 
 \tabularnewline
 & \textbf{P} & \textbf{R} & \textbf{F1}  & \textbf{P} & \textbf{R} & \textbf{F1} &  \textbf{P} & \textbf{R} & \textbf{F1}  \tabularnewline
\hline 
BW & 61 & 61 & 61 & 45 & 45 & 45 & 34 & 36 & 32\\
BC & 63 & 63 & \textbf{63} & 48 & 47 & \textbf{47} & 42 & 39 & 34\\ \hline
FT & 61 & 62 & 61 & 47 & 49 & 46 & 38 & 39 & \textbf{36}\\
\hline 
\end{tabular}
}
\caption{\label{tab:comparison} Comparison of B-LSTM with word modeling (BW), B-LSTM with character modeling (BC), and FastText (FT) on the same Twitter emoji prediction tasks proposed by \citeauthor{barbieri2017emojis} (\citeyear{barbieri2017emojis}), using the same Twitter dataset.}
\end{table}

To compare the FastText model with the word and character based B-LSTMs presented by \newcite{barbieri2017emojis}, we consider the same three emoji prediction tasks they proposed: top-5, top-10 and top-20 emojis most frequently used in their Tweet datasets. In this comparison we used the same Twitter datasets.
As we can see in Table~\ref{tab:comparison} FastText model is competitive, and it is also able to outperform the character based B-LSTM in one of the emoji prediction tasks (top-20 emojis). This result suggests that we can employ FastText to represent Social Media short text (such as Twitter or Instragram) with reasonable accuracy.

\begin{table}[ht!]
\centering
\setlength{\tabcolsep}{3pt} 
\renewcommand{\arraystretch}{1} 
\scalebox{0.85}{
\begin{tabular}{|r|ccc|ccc|ccc|}
\hline 
 & \multicolumn{3}{c|}{\textbf{top-5}} 
 & \multicolumn{3}{c|}{\textbf{top-10}}
 & \multicolumn{3}{c|}{\textbf{top-20}} 
 \tabularnewline
 & \textbf{P} & \textbf{R} & \textbf{F1}  & \textbf{P} & \textbf{R} & \textbf{F1} &  \textbf{P} & \textbf{R} & \textbf{F1}  \tabularnewline
\hline 
Maj & 7.9 & 20.0 & 11.3 & 2.7 & 10.0 & 4.2 & 0.9 & 5.0 & 1.5 \\ 
W.R. & 20.1 & 20.0 & 20.1 & 9.8 & 9.8 & 9.8 & 4.6 & 4.8 & 4.7 \\ \hline
\textbf{Vis} & 38.6 & 31.1 & 31.0 & 26.3 & 20.9 & 20.5 & 20.3 & 17.5 & 16.1 \\
\textbf{Tex} & 56.1 & 54.4 & 54.9 & 41.6 & 37.5 & 38.3 & 36.7 & 29.9 & 31.3 \\
\textbf{Mul} & 57.4 & 56.3 & \textbf{56.7} & 42.3 & 40.5 & \textbf{41.1} & 36.6 & 35.2 & \textbf{35.5} \\ \hline
\textit{\%} & 2.3 & 3.5 & 3.3 & 1.7 & 8 & 7.3 & -0.3 & 17.7 & 13.4 \\
\hline 
\end{tabular}
}
\caption{\label{tab:results} Prediction results of top-5, top-10 and top-20 most frequent emojis in the Instagram dataset: Precision (P), Recall (R), F-measure (F1). Experimental settings: majority baseline, weighted random, visual, textual and multimodal systems. In the last line we report the percentage improvement of the multimodal over the textual system.}
\end{table}

\subsection{Multimodal Emoji Prediction}
\label{sec:mm}
We present the results of the three emoji classification tasks, using  the visual, textual and multimodal features (see Table~\ref{tab:results}). 

The emoji prediction task seems difficult by just using the image of the Instagram post (\textbf{Visual}), even if it largely outperforms the majority baseline\footnote{Always predict \includegraphics[height=0.32cm,width=0.32cm]{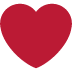} since it is the most frequent emoji.} and weighted random\footnote{Random keeping labels distribution of the training set}. 
We achieve better performances when we use feature embeddings extracted from the text. 
The most interesting finding is that when we use a multimodal combination of visual and textual features, we get a non-negligible improvement. This suggests that these two modalities embed different representations of the posts, and when used in combination they are synergistic. It is also interesting to note that the more emojis to predict, the higher improvement the multimodal system provides over the text only system (3.28\% for top-5 emojis, 7.31\% for top-10 emojis, and 13.42 for the top-20 emojis task). 

\subsection{Qualitative Analysis}
\label{sec:learn} 
In Table~\ref{tab:20detailed} we show the results for each class in the top-20 emojis task. 

The emoji with highest F1 using the textual features is the most frequent one \includegraphics[height=0.32cm,width=0.32cm]{emo/2764} (0.62) and the US flag 
\includegraphics[height=0.32cm,width=0.32cm]{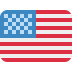} (0.52). The latter seems easy to predict since it appears in specific contexts: when the word USA/America is used (or when American cities are referred, like \#NYC).

The hardest emojis to predict by the text only system are the two gestures
\includegraphics[height=0.32cm,width=0.32cm]{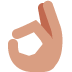} (0.12) and
\includegraphics[height=0.32cm,width=0.32cm]{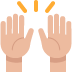} (0.13). The first one is often selected when the gold standard emoji is the second one or
\includegraphics[height=0.32cm,width=0.32cm]{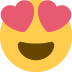}
\includegraphics[height=0.32cm,width=0.32cm]{emo/1f64c} is often mispredicted by wrongly selecting
\includegraphics[height=0.32cm,width=0.32cm]{emo/1f60d} or
\includegraphics[height=0.32cm,width=0.32cm]{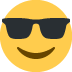}. 

Another relevant confusion scenario related to emoji prediction has been spotted by \newcite{barbieri2017emojis}: relying on Twitter textual data they showed that the emoji \includegraphics[height=0.32cm,width=0.32cm]{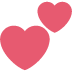} was hard to predict as it was used similarly to \includegraphics[height=0.32cm,width=0.32cm]{emo/2764}. Instead when we consider Instagram data, the emoji \includegraphics[height=0.32cm,width=0.32cm]{emo/1f495} is easier to predict (0.23), even if it is often confused with 
\includegraphics[height=0.32cm,width=0.32cm]{emo/1f60d}.

When we rely on visual contents (Instagram picture), the emojis which are easily predicted are the ones in which the associated photos are similar. For instance, most of the pictures associated to \includegraphics[height=0.32cm,width=0.32cm]{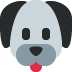}
are dog/pet pictures. Similarly, \includegraphics[height=0.32cm,width=0.32cm]{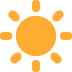} is predicted along with very bright pictures taken outside. \includegraphics[height=0.32cm,width=0.32cm]{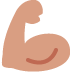} is correctly predicted along with pictures related to gym and fitness.
The accuracy of \includegraphics[height=0.32cm,width=0.32cm]{emo/1f4aa} is also high since most posts including this emoji are related to fitness (and the pictures are simply either selfies at the gym, weight lifting images, or protein food).

\begin{figure}
\centering
\includegraphics[height=7.5cm,keepaspectratio]{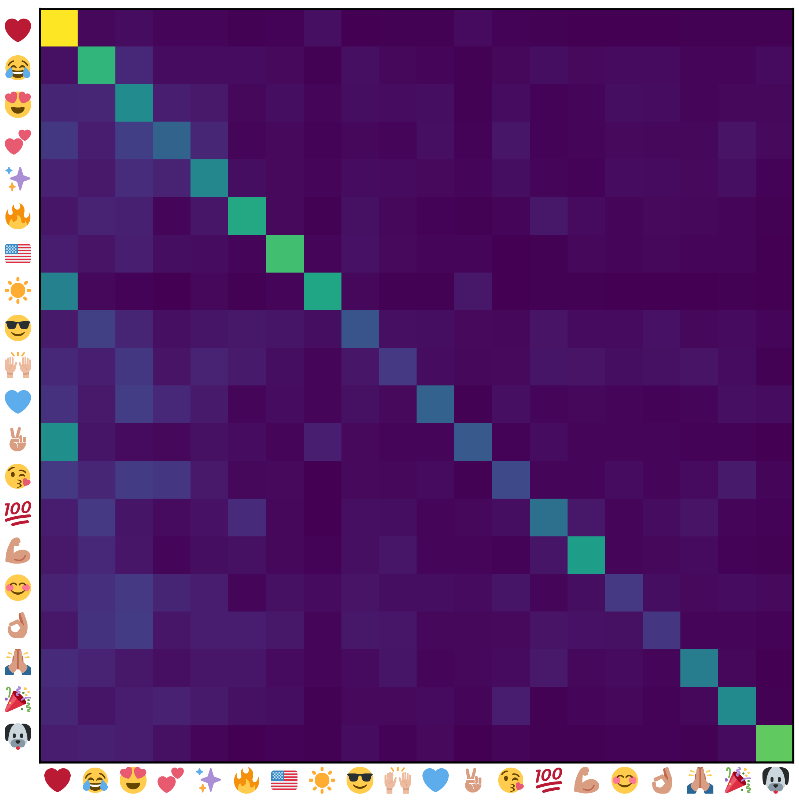} \\ 
\caption{Confusion matrix of the multimodal model. The gold labels are plotted as y-axes and the predicted labels as x-axes. The matrix is normalized by rows.}
\label{fig:cm}
\end{figure}

Employing a multimodal approach improves performance. This means that the two modalities are somehow complementary, and adding visual information helps to solve potential ambiguities that arise when relying only on textual content. In Figure~\ref{fig:cm} we report the confusion matrix of the multimodal model. The emojis are plotted from the most frequent to the least, and we can see that the model tends to mispredict emojis selecting more frequent emojis (the left part of the matrix is brighter).

\begin{table}
\centering
\setlength{\tabcolsep}{3.3pt} 
\scalebox{0.85}{
\begin{tabular}{|cc|ccc||cc|ccc|} \hline
\textbf{E} & \textbf{\%} & \textbf{Tex} & \textbf{Vis} & \textbf{MM} & \textbf{E} & \textbf{\%} & \textbf{Tex} & \textbf{Vis} & \textbf{MM}   \\ \hline
\includegraphics[height=0.42cm,width=0.42cm]{emo/2764} & 17.46 & 0.62 & 0.35 & \bf0.69 & \includegraphics[height=0.42cm,width=0.42cm]{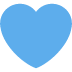} & 3.68 & 0.22 & 0.15 & \bf0.29 \\
\includegraphics[height=0.42cm,width=0.42cm]{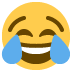} & 9.10 & 0.45 & 0.30 & \bf0.47 & \includegraphics[height=0.42cm,width=0.42cm]{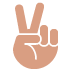} & 3.55 & 0.20 & 0.02 & \bf0.26 \\
\includegraphics[height=0.42cm,width=0.42cm]{emo/1f60d} & 8.41 & 0.32 & 0.15 & \bf0.34 & \includegraphics[height=0.42cm,width=0.42cm]{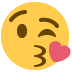} & 3.54 & 0.13 & 0.02 & \bf0.2 \\
\includegraphics[height=0.42cm,width=0.42cm]{emo/1f495} & 5.91 & 0.23 & 0.08 & \bf0.26 & \includegraphics[height=0.42cm,width=0.42cm]{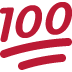} & 3.51 & 0.26 & 0.17 & \bf0.31 \\
\includegraphics[height=0.42cm,width=0.42cm]{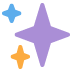} & 5.73 & 0.35 & 0.17 & \bf0.36 & \includegraphics[height=0.42cm,width=0.42cm]{emo/1f4aa} & 3.31 & 0.43 & 0.25 & \bf0.45 \\
\includegraphics[height=0.42cm,width=0.42cm]{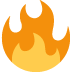} & 4.58 & 0.45 & 0.24 & \bf0.46 & \includegraphics[height=0.42cm,width=0.42cm]{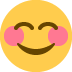} & 3.25 & 0.12 & 0.01 & \bf0.16 \\
\includegraphics[height=0.42cm,width=0.42cm]{emo/1f1fa-1f1f8} & 4.31 & 0.52 & 0.23 & \bf0.53 & \includegraphics[height=0.42cm,width=0.42cm]{emo/1f44c} & 3.14 & 0.12 & 0.02 & \bf0.15 \\
\includegraphics[height=0.42cm,width=0.42cm]{emo/2600} & 4.15 & 0.38 & 0.26 & \bf0.49 & \includegraphics[height=0.42cm,width=0.42cm]{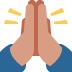} & 3.11 & 0.34 & 0.11 & \bf0.36 \\
\includegraphics[height=0.42cm,width=0.42cm]{emo/1f60e} & 3.84 & 0.19 & 0.1 & \bf0.22 & \includegraphics[height=0.42cm,width=0.42cm]{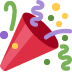} & 2.91 & 0.36 & 0.04 & \bf0.37 \\
\includegraphics[height=0.42cm,width=0.42cm]{emo/1f64c} & 3.73 & 0.13 & 0.03 & \bf0.16 & \includegraphics[height=0.42cm,width=0.42cm]{emo/1f436} & 2.82 & 0.45 & 0.54 & \bf0.59 \\
\hline
\end{tabular}
}
\caption{\label{tab:20detailed}F-measure in the test set of the 20 most frequent emojis using the three different models. ``\%'' indicates the percentage of the class in the test set}
\end{table}

\subsubsection{Saliency Maps}
\begin{figure}
\centering
\includegraphics[height=7cm,keepaspectratio]{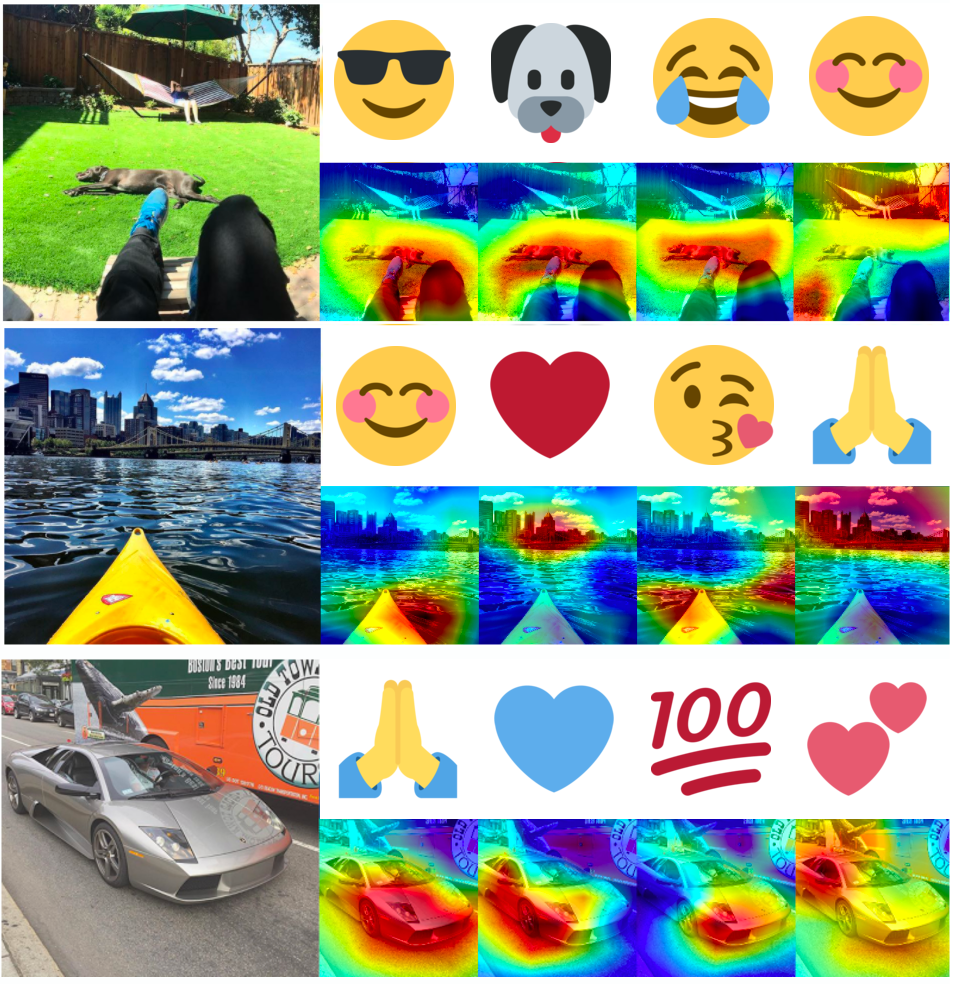} \\ 
\caption{Three test pictures. From left to right, we show the four most likely predicted emojis and their correspondent class activation mapping heatmap.}
\label{fig:imgFoc_ex1}
\end{figure}

In order to show the parts of the image most relevant for each class we analyze the global average pooling \cite{lin2013network} on the convolutional feature maps \cite{zhou2016learning}.
By visually observing the image heatmaps of the set of Instagram post pictures we note that in most cases it is quite difficult to determine a clear association between the emoji used by the user and some particular portion of the image. Detecting the correct emoji given an image is harder than a simple object recognition task, as the emoji choice depends on subjective emotions of the user who posted the image. In Figure \ref{fig:imgFoc_ex1} we show the first four predictions of the CNN for three pictures, and where the network focuses (in red). We can see that in the first example the network selects the smile with sunglasses \includegraphics[height=0.32cm,width=0.32cm]{emo/1f60e} because of the legs in the bottom of the image, the dog emoji \includegraphics[height=0.32cm,width=0.32cm]{emo/1f436} is selected while focusing on the dog in the image, and the smiling emoji \includegraphics[height=0.32cm,width=0.32cm]{emo/1f60a} while focusing on the person in the back, who is lying on a hammock. In the second example the network selects again the \includegraphics[height=0.32cm,width=0.32cm]{emo/1f60a} due to the water and part of the kayak, the heart emoji \includegraphics[height=0.32cm,width=0.32cm]{emo/2764} focusing on the city landscape, and the praying emoji \includegraphics[height=0.32cm,width=0.32cm]{emo/1f64f} focusing on the sky. The same ``praying'' emoji is also selected when focusing on the luxury car in the third example, probably because the same emoji is used to express desire, i.e. ``please, I want this awesome car''.

It is interesting to note that images can give context to textual messages like in the following Instagram posts: 
(1)``Love my new home \includegraphics[height=0.32cm,width=0.32cm]{emo/2600}'' (associated to a picture of a bright garden, outside) and (2) ``I can't believe it's the first day of school!!! I love being these boys' mommy!!!! $\#myboys$ $\#mommy$ \includegraphics[height=0.32cm,width=0.32cm]{emo/1f499}'' (associated to picture of two boys wearing two blue shirts).
In both examples the textual system predicts \includegraphics[height=0.32cm,width=0.32cm]{emo/2764}. While the multimodal system correctly predicts both of them: the blue color in the picture associated to (2) helps to change the color of the heart, and the sunny/bright picture of the garden in (1) helps to correctly predict \includegraphics[height=0.32cm,width=0.32cm]{emo/2600}.

\section{Related Work}
Modeling the semantics of emojis, and their applications, is a relatively novel research problem with direct applications in any social media task. Since emojis do not have a clear grammar, it is not clear their role in text messages. Emojis are considered function words or even affective markers \citep{na2017varying}, that can potentially affect the overall semantics of a message \citep{donato2017investigating}.

Emojis can encode different meanings, and they can be interpreted differently. Emoji interpretation has been explored user-wise \citep{miller2017understanding}, location-wise, specifically in countries \citep{barbieri2016cosmopolitan} and cities \citep{espinosa2016revealing}, and gender-wise \citep{chen2017through} and time-wise \citep{barbieri2018time}.

Emoji sematics and usage have been studied with distributional semantics, with models trained on Twitter data \citep{barbieri2016does}, Twitter data together with the official unicode description \citep{eisner2016emoji2vec}, or using text from a popular keyboard app
\newcite{ai2017untangling}. In the same context, \newcite{wijeratne2017emojinet} propose a platform for exploring emoji semantics. In order to further study emoji semantics, two datasets with pairwise emoji similarity, with human annotations, have been proposed: EmoTwi50 \citep{barbieri2016does} and EmoSim508 \citep{wijeratne2017semantics}. Emoji similarity has been also used for proposing efficient keyboard emoji organization  \citep{pohl2017beyond}.
Recently, \newcite{barbieri2018modifiers} show that emoji modifiers (skin tones and gender) can affect the semantics vector representation of emojis.

Emoji play an important role in the emotional content of a message. 
Several sentiment lexicons for emojis have been proposed \citep{novak2015sentiment,kimura2017automatic,rodrigues2018lisbon} and also studies in the context of emotion and emojis have been published recently \citep{wood2016emoji,hu2017spice}. 

During the last decade several studies have shown how sentiment analysis improves when we jointly leverage information coming from different modalities (e.g. text, images, audio, video)  \cite{morency2011towards, poria2015deep, tran2018ensemble}. In particular, when we deal with Social Media posts, the presence of both textual and visual content has promoted a number of investigations 
on sentiment or emotions 
\cite{baecchi2016multimodal, you2016cross, you2016robust, yu2016visual, chen2015multimodal} or emojis \citep{cappallo2015image2emoji,cappallo2018new}.

\section{Conclusions}
\label{sec:conclusions}
In this work we explored the use of emojis in a multimodal context (Instagram posts). We have shown that using a synergistic approach, thus relying on both textual and visual contents of social media posts, we can outperform state of the art unimodal approaches (based only on textual contents).
As future work, we plan to extend our models by considering the prediction of more than one emoji per Social Media post and also considering a bigger number of labels. 

\section*{Acknowledgments}
We thank the anonymous reviewers for their important suggestions. Francesco B. and Horacio S. acknowledge support from the TUNER project (TIN2015-65308-C5-5-R, MINECO/FEDER, UE) and the Maria de Maeztu Units of Excellence Programme (MDM-2015-0502).

\bibliography{aaai2018biblio}
\bibliographystyle{acl_natbib}
\end{document}